\documentclass[default,iicol]{sn-jnl}%
\jyear{2021}%

\theoremstyle{thmstyleone}%
\theoremstyle{thmstyletwo}%

\theoremstyle{thmstylethree}%

\usepackage{amsmath}
\usepackage{graphicx}
\usepackage{subfigure}
\usepackage{url}
\usepackage[utf8]{inputenc}
\usepackage{algorithm}
\usepackage{algpseudocode}
\begin{document}

\title[faster-rcnn-retail]{An Improved Deep Learning Approach For Product Recognition on Racks in Retail Stores}

\author[1]{\fnm{Ankit} \sur{Sinha}}\email{ankitsinha.cse18@itbhu.ac.in}

\author[2]{\fnm{Soham} \sur{Banerjee}}\email{banerjeesoham524@gmail.com}

\author*[1]{\fnm{Pratik} \sur{Chattopadhyay}}\email{pratik.cse@iitbhu.ac.in}

\affil*[1]{\orgdiv{Pattern Recognition Laboratory, Department of Computer Science and Engineering}, \orgname{Indian Institute of Technology (BHU)},\orgaddress{ \city{Varanasi}, \postcode{221005}, \state{Uttar Pradesh}, \country{India}}}

\affil[2]{\orgdiv{Department of Computer Science and Engineering}, \orgname{Budge Budge Institute of Technology}, \orgaddress{ \city{Kolkata},  \state{West Bengal}, \country{India}}}



\abstract{
Automated product recognition in retail stores is an important real-world application in the domain of Computer Vision and Pattern Recognition. In this paper, we consider the problem of automatically identifying the classes of the products placed on racks in retail stores from an image of the rack and information about the query/product images. We improve upon the existing approaches in terms of effectiveness and memory requirement by developing a two-stage object detection and recognition pipeline comprising of a Faster-RCNN-based object localizer that detects the object regions in the rack image and a ResNet-18-based image encoder that classifies the detected regions into the appropriate classes. Each of the models is fine-tuned using appropriate data sets for better prediction and data augmentation is performed on each query image to prepare an extensive gallery set for fine-tuning the ResNet-18-based product recognition model. This encoder is trained using a triplet loss function following the strategy of online-hard-negative-mining for improved prediction. The proposed models are lightweight and can be connected in an end-to-end manner during deployment for automatically identifying each product object placed in a rack image. Extensive experiments using Grozi-32k and GP-180 data sets verify the effectiveness of the proposed model.} 




\maketitle

\section{Introduction}\label{intro}

\begin{figure*}[ht]
    \centering
    \includegraphics[width = 0.99\textwidth]{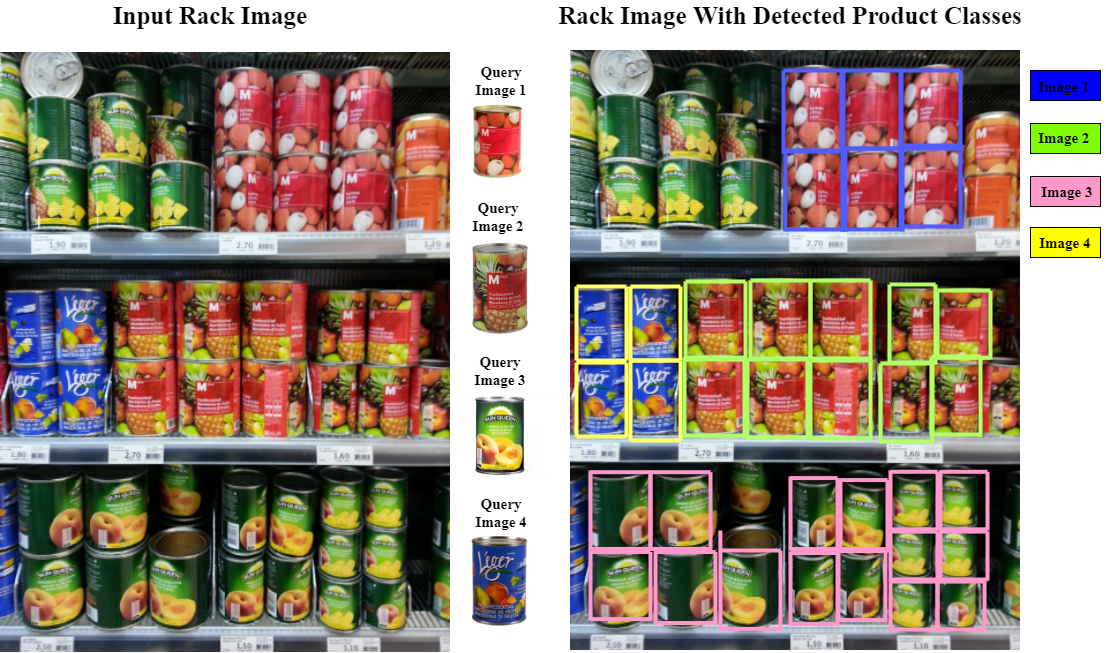}
    \caption{Problem scenario considered in this work}
    \label{fig:sample}
\end{figure*}

Computer Vision and Artificial Intelligence have an immense impact on the automation of processes in several industries including retail. Physical stores are getting gradually equipped with sensor arrays, cameras, integrated chips \cite{CV-transform-retail}. In this paper, we propose a Deep Learning-based solution to the problem of identifying products on racks in retail stores from an image of the rack and a database of query/product images that may be placed on the rack. The problem is explained in further detail with the help of a rack image in Fig. \ref{fig:sample}.
With reference to the figure, we have a list of four query images, and multiple instances of each of these may be placed at different positions on the rack. Given a rack image, our task is to identify the precise location of each query item in the rack image. The image on the extreme left shows the input rack image, whereas the image on the extreme right shows the detected product instances in the rack image. In the figure, instances of \textit{Query Image 1} are marked with blue bounding boxes, whereas instances of \textit{Query Image 2}, \textit{Query Image 3}, and \textit{Query Image 4} are marked with green, pink, and yellow colored bounding boxes, respectively. In real-life situations, there will be several such query images and large rack images due to which the need for the development of automated product recognition algorithms in retail stores cannot be over-emphasized.

There are several challenges associated with the problem as given in \cite{best-review-cv-in-retail,santra-review-CVRetail}. 
Usually, only a single query/reference image per product class is available which is insufficient to train a Deep Neural Network model. Also, the appearance of the reference image may differ significantly from the rack images in terms of orientation, illumination, resolution, reflection, etc. 
Moreover, 
variants of the same product with minor visual cues are likely to pose difficulty in the correct identification of the actual product class. 

The problem has several use-cases each of which has a great scope for automation, and inventory management is one of them. In retail stores, some people are required to continuously keep track of the available product stocks and place orders to procure more quantity of the sold-out items. Attempts are underway to automate this process using cameras and Internet of Things (IoT) technology. Another use case is assisting customers in retail stores by precisely locating products of their choice from several items placed in a large number of racks. 
Planogram consistency check is another use case. 
Planogramming is the process of designing an ideal layout for putting items on store shelves to boost sales. Hence, verifying whether a planogram has been properly implemented is very crucial. This task requires manual efforts to move from rack to rack and physically examine if the products are placed complying with the planogram, which is time-intensive as well as cost-intensive. To date, several research work have tried to address this issue and reduce human intervention as in \cite{Unsup-plano-compl-paper,sub-graph-isomorph-gp180,planogram-recur,robust-vis-analysis-plano-comp, vision-based-chg-detect-plano-comp,korean-recur,santra-ERPCNN,santra-GNMS,A-Ton-DL-pipeline,M-George-paper}. 

Our work is an improvement to that proposed by Tonioni \textit{et al.} in \cite{A-Ton-DL-pipeline} that also follows a two-stage pipeline involving Deep Neural Networks for object localization and recognition. Specifically, we attempt to improve the effectiveness of the YOLO-based product localization model used in \cite{A-Ton-DL-pipeline} by employing a Faster-RCNN with a Feature Pyramid Network to capture multi-scale features for improved localization. For object recognition also, we use a lightweight ResNet-18-based product recognition model instead of the larger VGG-16 model as considered in  \cite{A-Ton-DL-pipeline}.
The underlying architecture of our proposed model is lightweight and is expected to perform more accurately due to the use of multi-scale features for product localization 
as compared to that of \cite{A-Ton-DL-pipeline}, which makes it suitable for deployment on edge devices with less amount of memory. 
The main contributions of this paper can be summarized as follows:
\begin{itemize}
\item In this article, a two-stage Deep Learning-based pipeline for product recognition in retail settings is proposed which is lightweight and can be conveniently integrated with edge devices. 
\item We avoid template matching-based object detection as used by most existing techniques and employ multi-scale Deep Convolutional Networks for the same task which makes the prediction accurate with a fast response time. This scheme is also capable of effectively identifying blank regions on the rack which the template matching-based approaches fail to handle properly.

\item We improve upon the work in \cite{A-Ton-DL-pipeline} in terms of employing a more effective network that extracts multi-scale features for product localization and also using a much lighter ResNet-18 model for product recognition.
\item We conduct extensive experiments and perform a comparative study with state-of-the-art approaches to evaluate the effectiveness of the proposed approach.
\end{itemize}

The remainder of this paper is organized as follows: Section \ref{related} highlights the related work done in this domain. Section \ref{proposed} elaborates the proposed approach and Section \ref{results} describes the experiments performed by our team to validate our approach and presents the results. Finally, 
Section \ref{conclusions} concludes the work and highlights the future scopes for research in this area.

\section{Related Work} \label{related}
 Santra \textit{et al.} have been working on several challenges in this domain \cite{santra-ERPCNN,santra-GNMS,santra-conv-lstm}. In \cite{santra-ERPCNN}, an end-to-end annotation-free mechanism for product detection on racks is proposed. It is a multi-stage  exemplar-driven approach in which the relative scale of the rack images with respect to the available product templates is estimated in the first stage. In the second stage, potential object regions are determined and refined using greedy Non-Max-Suppression (NMS). Finally, a Convolutional Neural Network (CNN) is used to perform the classification using the extracted regions to identify the product classes. In \cite{santra-GNMS}, the use of greedy NMS in the task of detecting object locations in rack images is analyzed in depth. In this work, the authors argue that the greedy NMS discards bounding boxes with superior geometric placement due to overlapping of other boxes with higher confidence scores and propose a graph-based NMS to compute the potential confidence scores. 
 In another work \cite{santra-conv-lstm}, fine-grained classification of product instances is done by extracting unique local patches around key points within an image and encoding these using Convolutional Long-Short-Term Memory (LSTM) network. 

George \textit{et al.} \cite{M-George-paper} proposed a per-exemplar multi-label image classification and localization approach 
by establishing a locality constraint linear 
coding \cite{locality-const-linear-coding} model using dense SIFT features of the product images. Further, a discriminative Random Forest is trained followed by a multi-class ranking of products to carry out the recognition task. Wang \textit{et al.} \cite{SADCL} proposed a destruction followed by construction method guided by a self-attention mechanism for end-to-end fine-grained classification tasks. Osokin \textit{et al.} \cite{OS2D} employed a one-stage hybrid model to achieve the tasks of localization and recognition jointly. First, the local features are extracted from both the input and class images using a ResNet following which dense correlation scores between the two are computed. Next, the feature maps are semantically aligned through a trained geometric transformation model to predict the bounding boxes before finally computing the recognition scores.

In the task of planogram compliance, Saran \textit{et al.} \cite{robust-vis-analysis-plano-comp} presented a visual analysis framework and applied the Hausdorff metric to compute occupancy of product shelves. Here, the authors describe a robust product counting algorithm using row detection methods with texture and color-based features. In another work, Ray \textit{et al.} \cite{Unsup-plano-compl-paper} proposed a two-layer hypothesis and verification model in which the model first predicts a set of candidate items at a certain position of the rack, and next the above hypothesis is verified by a graph-based algorithm. The proposal about candidate items is made through a combination of correlation-based and ad-hoc SURF \cite{SURF-paper} schemes.
Liu \textit{et al.} \cite{planogram-recur} proposed an unsupervised recurrent pattern mining strategy with a graph-based matching algorithm for planogram compliance by adopting a divide and conquer policy independent of product templates. In \cite{sub-graph-isomorph-gp180}, the authors proposed a two-stage approach to recognize products in rack images. First, SIFT features and Hough transformation are used to find probable matches of reference product images on the shelf. Next, to determine the missing products and remove incorrect matches, sub-graph isomorphism between the observed output and the actual output is performed.

Goldman and Goldberger \cite{CRF-context} proposed a Deep Learning-based method to classify well-structured objects with high inter-class similarity by treating sequences of images as linear Conditional Random Fields (CRFs) to include contextual information. Baz \textit{et al.} \cite{cntxt-aware-hyb-sys-fine-grained-prod-recog} proposed a hybrid classifier combining Support Vector Machines (SVM) with probabilistic graphical models like Hidden Markov Models (HMM) and CRFs by exploiting the spatial continuity in the arrangement of products on a rank. Since products of similar brands are placed adjacent to each other, the authors model the contextual information using an \textit{HMM} or \textit{CRF}. The classification performance of SVM has been seen to improve with the inclusion of this contextual information.
A method for fine-grained classification of products by detecting recurring features 
in rack images is presented in \cite{korean-recur}. These recurring features are compared with SIFT features from the logo regions of the reference images and assigned a rough class label. Further, fine-grained classification has been performed by training a VGG-16 \cite{VGG-paper} with an attention map policy generated using matching SURF \cite{SURF-paper} and BRISK \cite{BRISK-paper} features from the product instance and the template image of the rough class label. However, such a template matching-based approach is likely to be time-intensive due to the exhaustive search required across the entire rack image and is not suitable for most practical purposes. The work in
Tonioni \textit{et al.} \cite{A-Ton-DL-pipeline} is one of the few approaches that consider Deep Learning-based prediction for both the object localization and prediction. Here, the authors make use of a two-stage pipeline based on YOLO-v2 \cite{yolo-v2} and VGG-16 \cite{VGG-paper}, in which the YOLO-v2 object detector is fine-tuned using a privately annotated dataset to predict the bounding boxes of the product instances. Further, the VGG-16 embedder is trained with triplet loss \cite{facenet-paper} and MAC \cite{MAC-paper} features to identify the product class. 

From the extensive literature survey, it has been found that classical image processing and template matching-based methods dominate Deep Learning-based techniques in the majority of the retail use-cases. The scarcity of sufficient ground truth for product images is one of the primary causes for this. 
Template matching-based techniques 
suffer from high response time and are also susceptible to noise. On the other hand, Neural Network-based approaches are known for their robustness against noise and variations of input conditions, and the few existing approaches in this category use heavyweight models that are not suitable for implementation on edge devices. In this work we employ effective but lightweight Deep Neural Network architectures for both object localization and product detection steps and perform a rigorous analysis and comparative study with state-of-the-art approaches. 
Our approach 
is described in detail in Section \ref{proposed}. 

\section{Proposed Approach}\label{proposed}
We implement a two-stage pipeline for the object localization and the recognition tasks. In the first stage, a faster RCNN-based Deep Neural Network \cite{faster-rcnn-paper} is employed to output the bounding box coordinates of the region proposals through a regressor head. In the second stage, another CNN-based Deep Neural Network model is used to convert these region proposals into feature descriptors in a latent space. Unlike previous models like YOLO \cite{YOLO-v1, yolo-v2,YOLO-v3}, SSD \cite{SSD-paper} that combine the region proposal and the recognition stages in a single model, our proposed two-stage architecture considers two separate dedicated lightweight models for the two above-mentioned tasks (as explained in the following two sub-sections), which is expected to improve the effectiveness of the overall approach further. 

\subsection{\textit{Localizing Product Instances}} \label{sec:localizer}
As mentioned before, we use the faster-RCNN architecture with ResNet-50 \cite{resnet-paper} backbone to extract features for localizing product instances. This feature extractor is followed by a Feature Pyramid Network \cite{FPN-paper} that aggregates useful multi-scale features in a top-down direction. 
Next, there is a region proposal network (RPN) that
uses anchors of fixed sizes and aspect ratios to output high-quality region proposals. After Region-of-Interest (RoI) pooling and Non-Maximal-Suppression (NMS), 
there is a bounding box \textit{regressor} head. The \textit{classifier} head of standard RCNN architecture is discarded since our embedder serves the same objective. Then, we extract patches cropped from the input image according to the predicted bounding boxes. 

\begin{figure*}[ht]
    \centering
    \includegraphics[width = 0.9\linewidth]{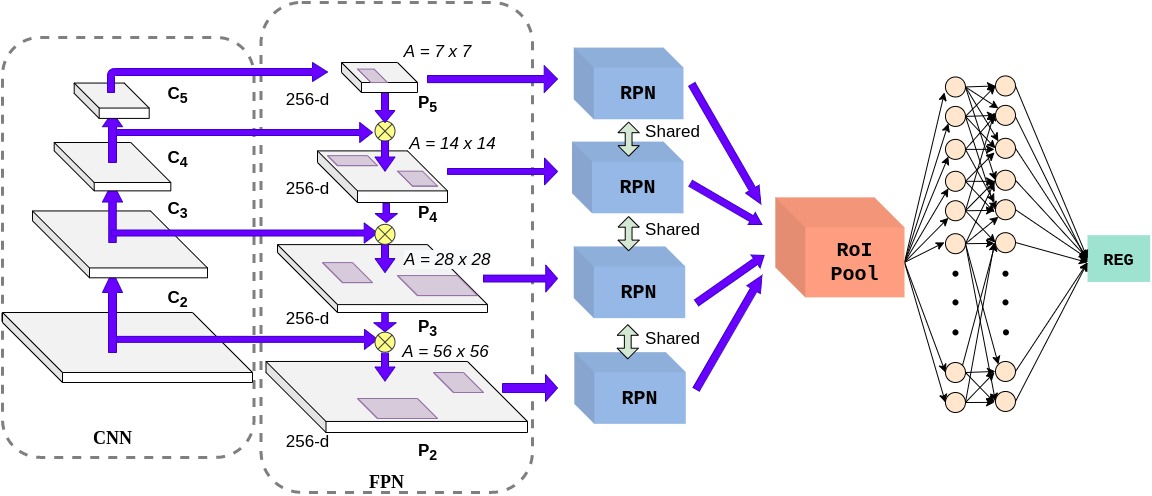}
    \caption{The Localization Pipeline} 
    \label{fig:localizer}
\end{figure*}

The complete architecture of the localization network is shown in Fig. \ref{fig:localizer}. 
The network consists of five convolutional blocks in which each block is a sub-network of multiple layers that produces feature maps of the same size. The outputs from the last layers of the second to fifth blocks 
are extracted and a reference set $\{C_2,C_3,C_4,C_5\}$ is formed. The feature map corresponding to the first block $C_1$ has not been included for further operations due to its large memory footprint. 
These feature maps are next used to construct a feature pyramid. The process of constructing the pyramid starts by passing $C_5$ through a $1 \times 1 \times d$ convolution layer to reduce the number of channels to $d$ and obtain the first pyramid feature $P_5$. Except for $P_5$, the rest of the feature pyramids are constructed by following a similar method. So, $P_5$ is upsampled to twice its size using nearest-neighbor interpolation. Let us call the output from this stage $\tilde{P}_5$. Next, $C_4$ is convolved using a $1 \times 1 \times d$ dimensional filter to obtain $\tilde{C}_4$ that has dimensions same as that of $\tilde{P}_5$. 
The feature maps $\tilde{C}_4$ and $\tilde{P}_5$ are henceforth added element-wise and this merged feature map is passed through another $3{\times}3$ dense convolution operation to compensate for the aliasing effect of upsampling and form the pyramid layer $P_4$. The depth of all the pyramid layers is set to $d = 256$. A similar procedure is also followed to obtain the feature pyramids $P_3$ and $P_2$. The construction of the feature pyramid $P_{k-1}$ from $P_k$ is mathematically expressed as follows: 
\begin{align}
    \tilde{P}_{k} \;\;\;\;&= \textrm{Upsample}_{\times 2}(P_{k}) \\
    \tilde{C}_{k-1} &= \textrm{Conv}_{1 \times 1}(C_{k-1}) \\
    P_{k-1} &= \textrm{Conv}_{3 \times 3}(\tilde{P}_{k} + \tilde{C}_{k-1})
\end{align}

The resulting feature pyramids carry a good balance of semantic and fine-grained information. So anchors of a specific scale are assigned to each pyramid level instead of repetitively assigning all scales of anchors to every level. This avoids computational redundancy and ensures the simplicity of design. The pyramid levels $P_2$, $P_3$, $P_4$, $P_5$ are assigned anchors of scales $32$, $64$, $128$, and $256$, respectively. Anchors of each scale have three aspect ratios: {1:1, 1:2, 2:1}. 

The FPN is followed by a region proposal network (RPN) \cite{faster-rcnn-paper} 
which is a mini convolutional network of 256 channels having $3{\times}3$ kernels. It further separates into two branches: \textit{object classifier} and \textit{object regressor}. 
The \textit{object classifier} branch performs \textit{object} vs \textit{background} binary classification  whereas the \textit{object regressor} predicts the bounding boxes for anchors with high objectness scores. Each of the feature pyramids, $P_5$, $P_4$, $P_3$, and $P_2$, is processed using an RPN in a sliding window manner. As shown in Fig. \ref{fig:localizer}, weight-sharing is done among all the RPNs 
and the outputs from the different RPNs are passed through a common Region-of-Interest (RoI) pooling layer. 
The RoIs are next passed through two-fully connected layers and the final bounding box \textit{regressor} head to localize the objects in the rack image. 

\subsection{\textit{Recognition of Products}}\label{sec:embedder}
\begin{figure}[ht]
    \centering
    \includegraphics[width = 0.95\linewidth]{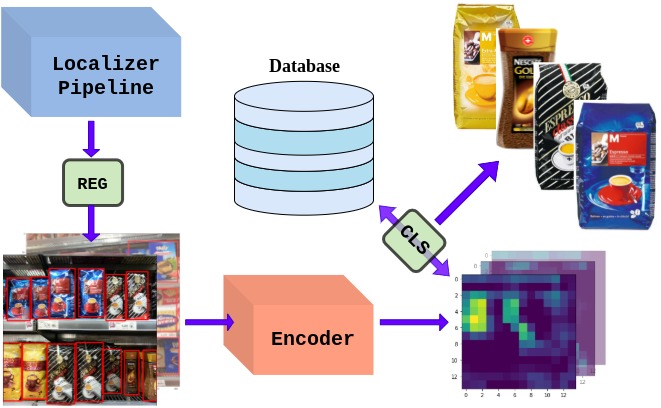}
    \caption{The Recognition Pipeline} 
    \label{fig:encoder}
\end{figure}

The recognition pipeline shown in Fig. \ref{fig:encoder} 
predicts the class of each object detected within the rack image in the previous stage.
A ResNet-18 embedder is used for this task which is trained offline by sampling triplets of different query images consisting of an anchor $i_{a}$, a positive  $i_{p}$, and a negative $i_{n}$ image. Let the embedding of these three images be denoted by $x_i^a$, $x_i^p$, and $x_i^n$ respectively. The anchor and the positive images belong to the same category while the negative image belongs to a different class. Using the common Euclidean distance function $d$ 
the network is trained to minimize the triplet loss $\mathcal{L}$ defined as:
\begin{align}
     \mathcal{L} = \frac{1}{N} \sum_{i=1}^N max(0, & d(x_i^a,x_i^p)  \!-\! d(x_i^a,x_i^n)\!+\!\alpha).
\end{align}
The objective here is to learn an embedding such that the following inequality is satisfied:
\begin{equation}
     \|x_i^a - x_i^p\|_2^2 + \alpha < \|x_i^a - x_i^n\|_2^2,
\end{equation}
where $\alpha$ denotes the fixed minimum margin between the distance pairs.
If a single image for every product class is present in the query image database, then we use data augmentation to generate multiple anchor images corresponding to each query image. For this, we apply standard image augmentation methods such as Gaussian blur, random crop, brightness, and saturation variations. A negative image for the triplet can be chosen in various ways. Theoretically, a huge number of triplets $(\approx N^2$) can be generated even from a small dataset, where \(N\) is the size of the 
dataset. Hence, to ensure faster convergence with good sample efficiency we apply the strategy of Online Hard Negative Mining (OHNM). 
In this approach, for every mini-batch of size \textit{b} loaded during training, the \textit{hardest} negative sample against each positive sample is chosen from the same batch. This selection scheme reduces the total computational complexity to $O(b N)$ from $O(N^2)$. 
Mathematically, 
\begin{eqnarray}
    X_{bat} \subset |D|,\\  
    x_i^a = \tilde{f}(x_i^p), \forall x_i^p \in X_{bat}, \label{eq7}\\
    \tilde{x}_i^n = argmin_{x_j}\|x_i^a - x_j\|_2^2, &\exists \tilde{x}_i^n \in X_{bat}.
\end{eqnarray}
In the above equations, $\tilde{f}$ denotes the image augmentation operator, $x_i^p$ denotes a positive sample, $x_i^n$ denotes a negative sample, $x_i^a$ denotes an anchor generated from $x_i^p$ using data augmentation, and $X_{bat}$ is a mini-batch of data. Through the minimization of this loss function, the network learns to encode images of the same class close to each other in the encoded space while separating those belonging to different classes. 

\begin{figure*}[ht]
    \centering
    \includegraphics[width = 0.9\linewidth]{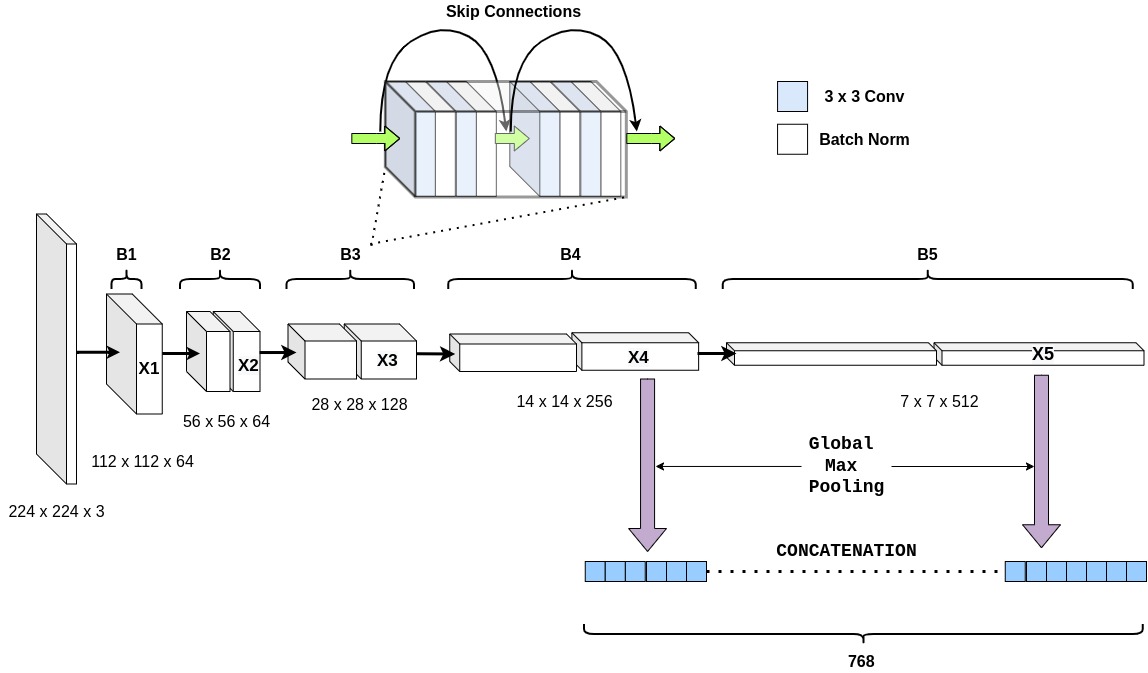}
    \caption{Block diagram of ResNet-18 based Encoder.}
    \label{fig:resnet-18}
\end{figure*}

The Embedder, as shown in Fig. \ref{fig:resnet-18}, is a ResNet-18 \cite{resnet-paper} pre-trained on the ImageNet1K \cite{imagenet-challenge} 
dataset. 
The ResNet-18 contains five convolutional blocks B1, B2, B3, B4, B5 that output feature maps of sizes 112, 56, 28, 14, and 7, respectively. The corresponding feature maps have been named \textit{X1}, \textit{X2}, \textit{X3}, \textit{X4}, and \textit{X5} in Fig. \ref{fig:resnet-18}. In our work, the final descriptors are extracted by applying MAC \cite{MAC-paper} operation on blocks B4 and B5, yielding $\tilde{X}_4$ and $\tilde{X}_5$, respectively, as shown below. 
\begin{align*}
\centering
    &X_4 = Conv4(X_3), && size = (14,14,256) \\
    &X_5 = Conv5(X_4), && size = (7,7,512) \\
    &\tilde{X_4} = GlobalMaxPool(X_4), && size = (1,256) \\
    &\tilde{X_5} = GlobalMaxPool(X_5), && size = (1,512) \\ 
    &X_{embed} = concat(\tilde{X_4},\tilde{X_5}), && size = (1,768)
\end{align*}
Next, the vectors are concatenated into a single descriptor and $\ell 2$ normalized. Mathematically,  
\begin{align}
\centering
    \tilde{X}_{embed} = \frac{X_{embed}}{\|X_{embed}\|_2}.
\end{align}

\section{Experiments}\label{results}
All experiments are conducted in Google Colab notebooks that provide free GPU and TPU support. PyTorch has been chosen as the deep learning framework. The colab virtual machines (VMs), by default, offer a RAM of 13 GB. The fine-tuning of Faster RCNN for object detection is done on a Nvidia Tesla T4 GPU with 16 GB memory. The embedder ResNet-18 is fine-tuned on a Nvidia Tesla K80 GPU with 12 GB capacity. 

The Faster-RCNN FPN, pretrained on COCO 2017 \cite{COCO-dataset}, is fine-tuned by Stochastic Gradient Descent (SGD) \cite{SGD-ref} with momentum $0.8$ and weight decay $5 \times 10^{-2}$ using a manually annotated small subset of 60 rack images from the Grozi-3.2K Food data. A learning rate schedule with warmup has been used for fine-tuning, where the number of warmup iterations is set to the total number of mini-batches $N_{batch} = \frac{\|D\|}{b}$. Here, $D$ denotes the dataset for fine-tuning, and $b$ (= 8) is the batch size. The warmup factor is set to $10^{-3}$ and the maximum learning rate is set to $5 \times 10^{-2}$. Having fine-tuned the detector for 25 epochs with the above settings, the learning rate is reduced to $5 \times 10^{-3}$ for additional 15 epochs. During fine-tuning, we use the top 1000 region proposals. 
\\
The Embedder ResNet-18 is the pre-trained checkpoint on ImageNet1K \cite{imagenet-challenge} dataset. It is fine-tuned using the standard triplet loss with a margin of $1.0$ using each product present in the query image database. In this work, we use the Grozi-3.2k data \cite{M-George-paper} that contains a total of 3235 query images, one for each product item. The fine-tuning is continued for $15$ epochs with a fixed learning rate of $10^{-4}$ using Adaptive Gradient Descent (ADAM) \cite{ADAM-paper} optimizer. 
For evaluation, we use 680 rack images from five different retail stores present in the Grozi-3.2k data \cite{M-George-paper}. However, the ground truth annotations for this data set consist of identical products, that are adjacently placed, grouped under a single bounding box. Hence, this data set is ideal for multi-label image classification tasks. We also use the GP-180 data \cite{sub-graph-isomorph-gp180} for evaluation which is a subset of the Grozi-3.2k data and contains instance-level annotations of 74 rack images. 

\begin{figure*}[ht]
\centering
    {
    \begin{subfigure}[ ]
    {\includegraphics[width = 0.30\textwidth]{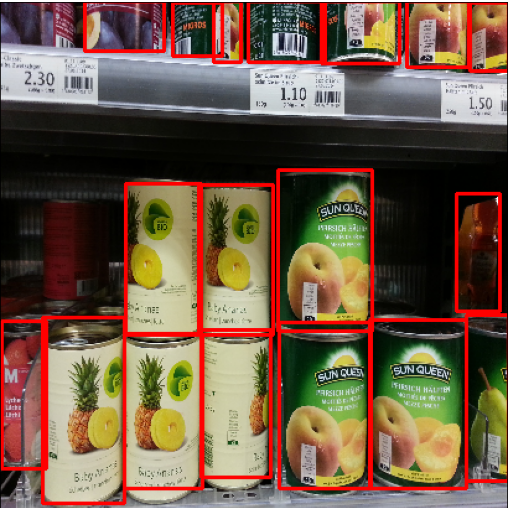}}
    \end{subfigure}
    }
    {
    \begin{subfigure}[ ]
    {\includegraphics[width = 0.30\textwidth]{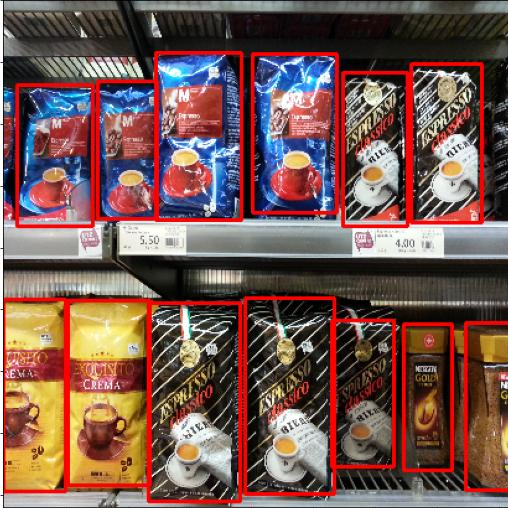}}
    \end{subfigure}
    }
    {
    \begin{subfigure}[ ]
    {\includegraphics[width = 0.30\textwidth]{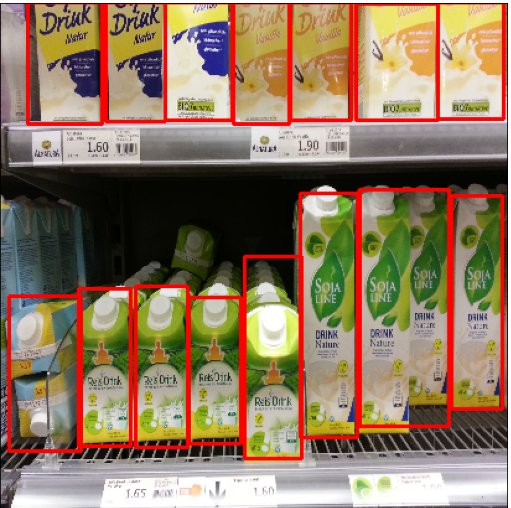}}
\end{subfigure}}
    \caption{Sample rack images from GP-180 and the bounding boxes predicted by our Faster-RCNN-based product detector marked in red}
\label{fig:rack-pred-pos}
\end{figure*}

In our first experiment, we evaluate the effectiveness of the product detection model (localizer) based on Faster-RCNN (refer to Section \ref{sec:localizer}). Fig. \ref{fig:rack-pred-pos} shows three rack images from GP-180 along with the predicted bounding boxes. It can be visually observed from the results that our detector accurately localizes the different items for each of the test images and it is also able to correctly identify the blank spaces (i.e., the regions on the rack with no objects). 

\begin{table}[ht]
    \caption{Performance of our Localization pipeline on GP-180 in the standard COCO format. Here, maxDets represents the number of region proposals used during testing.}
    \label{table:coco-detector}
    \begin{center}
    \begin{tabular}{c|c|c}
        \textbf{Metric} & \textbf{maxDets} & \textbf{Result} \\
        \hline \hline
        \noindent AP @[IoU = 0.50] & 100 & 0.864\\
        \hline 
        AP @[IoU = 0.75] & 100 & 0.726\\
        \hline
        AP @[IoU = 0.50:0.95] & 100 & 0.594\\
        \hline
        AR @[IoU = 0.50:0.95] & 1 & 0.059 \\
        \hline
        AR @[IoU = 0.50:0.95] & 10 & 0.539\\
        \hline 
        AR @[IoU = 0.50:0.95] & 100 & 0.672\\
        \hline
    \end{tabular}
\end{center}
\end{table}
We also study the effectiveness of our Faster RCNN-based object localization model using standard COCO metrics in Table \ref{table:coco-detector}. The GP-180 data has been used for this experiment. In the table, \textit{AP} denotes Average Precision, \textit{AR} denotes Average Recall, and \textit{maxDets} signifies the number of top region proposals considered during the test time. It is observed that our detection model is capable of precisely localizing the grocery products. For IoU in the range 0.50 to 0.95 (incremented in step-size of 0.05), the AP is $\sim 60 \%$. As far as the AR is concerned, it increases with the number of region proposals, as expected. However, the larger the number of region proposals, longer will be the inference time and lesser will be the AP. In a real-time use case (using only 10 region proposals), our model accurately localizes 54\% of the items on the rack, whereas on 100 region proposals AR satisfactorily increases to 67.2\%.

Each of the following experiments deals with the evaluation of our overall product detection and recognition pipeline and comparative study with other state-of-the-art techniques. First, we compare our work with \cite{korean-recur} and \cite{A-Ton-DL-pipeline} using the GP-180 \cite{sub-graph-isomorph-gp180} data. While the work in \cite{A-Ton-DL-pipeline} describes an end-to-end Deep Learning approach, that in \cite{korean-recur} is a hybrid method with template matching-based bounding box detector and a class-specific CNN recognizer. For this experiment, we follow the same protocol as described in \cite{sub-graph-isomorph-gp180}, i.e., we consider a prediction to be correct if the output label matches the ground truth label of the product, provided the IoU between the detected and ground-truth boxes is greater than 0.5. In Table \ref{table:dl-pipe-gp180}, we report the mean average precision mAP@0.5 and product recall PR@0.5 of our method, \cite{korean-recur}, and \cite{A-Ton-DL-pipeline}. 
\begin{table}[ht]
    \caption{Comparative results of our full pipeline on GP-180 according to the protocol adopted in \cite{A-Ton-DL-pipeline} and \cite{korean-recur}}
    \label{table:dl-pipe-gp180}
\begin{center}
    \begin{tabular}{c|c|c}
        \textbf{Method} & \textbf{mAP@0.5} & \textbf{PR@0.5} \\
        \hline \hline
        yolo\_ld+lf-mc-th \cite{A-Ton-DL-pipeline}& 76.93 & 86.56\\
        \hline 
        SIFT + vgg16 \cite{korean-recur}& \textbf{85.79} & -\\
        \hline
        frcnn + res50 + res18 (ours) & 82.70 & \textbf{89.70} \\
        \hline
    \end{tabular}
\end{center}
\end{table}
It is observed that the proposed method improves over \cite{A-Ton-DL-pipeline}  by $\sim 5$\% and  $\sim 3$\% in terms of mAP@0.5 and PR@0.5, respectively. Although \cite{korean-recur} shows $\sim 3$\% higher mAP@0.5 compared to that of ours, it involves the use of hand-crafted feature engineering for product localization which is significantly time-intensive and is expected to be less robust to the variation of input conditions. 

We also compare our approach with two recent approaches, namely \cite{santra-GNMS} and \cite{santra-ERPCNN}, on the GP-180 data set using a similar test protocol as specified in \cite{santra-GNMS, santra-ERPCNN}. Both these compared methods are hybrid approaches that use exemplar-driven localization and CNN-based object recognition framework. 
The testing protocol can be described as follows: 
Let a product \textit{P} be present on the rack. If the center of any detected bounding box lies within \textit{P} in the rack and the enclosed item is predicted as \textit{P}, the count of \textit{true-positives} (TP) is incremented by 1. If the center of the detected bounding box lies within \textit{P} in the rack but the enclosed item is not recognized as \textit{P} by the object recognition model, then the count of \textit{false-positives} (FP) of the rack is incremented by 1. Further, if the center of a detected box does not lie within any true product in the rack, the count FP is again increased by 1. Lastly, if there exists no detected box whose center lies within \textit{P}, the count of \textit{false-negatives} (FN) of the rack is incremented by 1. Considering the above, we compare our approach with that of \cite{santra-ERPCNN} and \cite{santra-GNMS} and present the $F_1$ scores after the final recognition phase in Table 
\ref{table:ISI}. 
\begin{table}[ht]
    \caption{Comparative results of our full pipeline on GP-180 dataset according to the evaluation protocol followed in \cite{santra-ERPCNN,santra-GNMS}.}
    \label{table:ISI}
\begin{center}
    \begin{tabular}{c|c}
        \textbf{Method} & \textbf{$F_1$ score (\%)} \\
        \hline \hline 
        ERP-CNN \cite{santra-ERPCNN}& 81.05 \\
        \hline 
        R-CNN-G \cite{santra-GNMS}& 80.21 \\
        \hline
        Ours & \textbf{83.20} \\
        \hline
    \end{tabular}
\end{center}
\end{table}
It can be seen from the table that our approach outperforms both the other two compared methods in terms of $F_1$ score by at least 2\%.

In the next experiment, we compare our approach with state-of-the-art methods  \cite{hough-trfm-obj-pose-recog,A-Ton-DL-pipeline,M-George-paper} using the Grozi-3.2k Food data set. Here, \cite{hough-trfm-obj-pose-recog} is a non-ML approach based on Hough transform, whereas \cite{M-George-paper} involves a fusion of several strategies including Fast dense pixel matching, Random Forests, and Genetic Algorithms. On the other hand, \cite{A-Ton-DL-pipeline} follows a fully Deep Learning pipeline. 
The evaluation protocol followed here is similar to that described in \cite{M-George-paper}. 
Specifically, we use two metrics, namely, mean average precision (mAP) and mean average product recall (mAPR). 
Corresponding to each bounding box detected by the localization model, we fetch the top $K$ predictions (candidate product items) by the object recognition model. If the ground truth label is present within these top $K$ predictions, the \textit{true-positive} count is increased by 1. Otherwise, the \textit{false-positive} count is incremented by 1. Next, the precision and product recall are computed and averaged across all the test images to obtain the average precision (AP) and average product recall (APR) scores. The AP and APR values computed for different values of $K$ are averaged to report the final mAP and mAPR metrics. 
The corresponding results are shown in Table \ref{table:Grozi-3.2k} for the compared methods for two values of $K$, i.e., 20 and 50. 
\begin{table}[ht]
    \caption{Comparative results on the Grozi-3.2K Food dataset using the protocol as described in \cite{M-George-paper}}
    \label{table:Grozi-3.2k}
\begin{center}
    \begin{tabular}{c|c|c}
        \textbf{Method} & \textbf{mAP (\%)} & \textbf{mAPR (\%)}\\
        \hline \hline 
        FM+HO \cite{hough-trfm-obj-pose-recog} & 23.71 & 41.60 \\
        \hline 
        yolo\_ld+lf-mc-th \cite{A-Ton-DL-pipeline}& 36.02 & \textbf{58.41} \\
        \hline
        RF+PM+GA \cite{M-George-paper} & 23.49 & 43.13 \\
        \hline
        Ours & \textbf{47.77} & 58.11 \\
        \hline
    \end{tabular}
\end{center}
\end{table}
It can be seen from the results that our method outperforms each of \cite{hough-trfm-obj-pose-recog, M-George-paper} both in terms of mAP and mAPR. It also surpasses the mAP score provided by \cite{A-Ton-DL-pipeline} by a large margin of 11\%, but in terms of mAPR we fall short of \cite{A-Ton-DL-pipeline} by only $\sim 0.30\%$. However, as also mentioned in Section \ref{intro}, the method in \cite{A-Ton-DL-pipeline} uses VGG-16 for object recognition which has a larger memory footprint compared to our ResNet-18-based product recognition model. While the YOLO-based object localization model has 62M parameters and VGG-16-based product recognition model has 138M parameters, our Faster RCNN-FPN-based localization model has only 42M parameters and the ResNet-18-based recognition model has only 11M parameters.  Hence, in terms of mAP, mAPR, and memory space requirement, our approach can be regarded as the best among the other competing methods used in this study.\\
    

\textbf{Shortcomings of our Approach}: 
\begin{figure*}[ht]
\centering
    {
    \begin{subfigure}[ ]
    {\includegraphics[width = 0.35\textwidth]{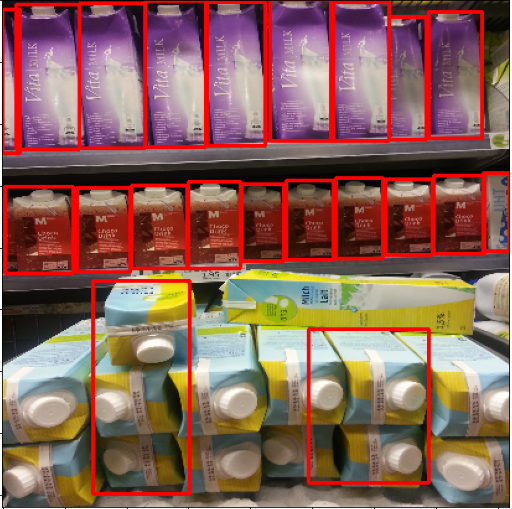}}
    \end{subfigure}
    }
    {
    \begin{subfigure}[ ]
    {\includegraphics[width = 0.35\textwidth]{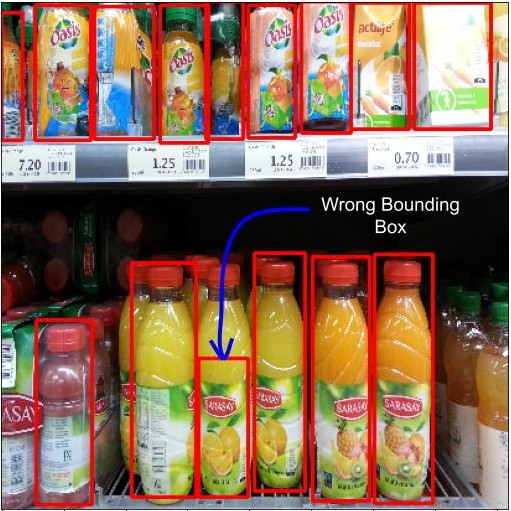}}
\end{subfigure}}
    \caption{Sample rack images from GP-180 and the bounding boxes predicted by our Faster-RCNN-based product detector marked in red.}
\label{fig:rack-pred-neg}
\end{figure*}
Despite the good performance, there are some potential areas of extension of our work, as highlighted next. For example, our product detection approach as well as other existing related methods fail to perform well if several items with similar appearances are stacked together one on top of the other in the rack. The scenario is explained using Fig. \ref{fig:rack-pred-neg}(a) in which the bottom shelf shows the top view of similar-looking boxes stacked together. It can be seen that the object regions predicted by our localization model (i.e., the red-colored boxes) corresponding to this part of the rack are not accurate enough. A possible way to counter this problem is to make the data set for training the Faster RCNN-based localization model more extensive by incorporating similar rack images and corresponding ground truth. 
Secondly, the proposed localization model tends to generate false positive boxes around objects with sharp texture variation in packaging as depicted in Fig. \ref{fig:rack-pred-neg}(b) which may cause the second classification stage to suffer. As can be seen from the figure, a sticker placed on a bottle has been incorrectly predicted as a distinct item. It appears that this problem can also be solved by adding related items in the gallery set for training the object localization model, which needs further study.



\section{Conclusions and Future Work}\label{conclusions}
In this work, we present a two-stage Deep Learning-based pipeline to automatically detect product locations and  identify the products placed on the racks in retail stores. The first stage involves the generation and refinement of region proposals using a Faster-RCNN-FPN model. In the second stage, these region proposals are passed through a ResNet-18-based embedder followed by a classification layer to predict the product class. During deployment, these two models are connected in an end-to-end manner to automatically identify which products are present on the shelves of a rack using only an image of the complete rack. We have made a thorough evaluation of both the localization and the recognition frameworks through extensive experiments and comparative study to verify the effectiveness of our approach.  

Due to the use of light-weight neural network-based models for both the localization and recognition phases, our approach is time-efficient and requires only a small amount of memory to run making it suitable for deployment on edge devices. Only a few existing approaches used in the comparative study, namely \cite{korean-recur,A-Ton-DL-pipeline}, show performance comparable to that of our model. However, these are either time-intensive due to the use of hand-crafted features for object detection or employ heavy-weight neural network models to perform the prediction, and are hence not suitable for large-scale applications. 
In contrast, our lightweight model can be conveniently used to keep track of inventories in large retail stores in a time-efficient manner. 
In the future, our model can be retrained with a more extensive data set to handle challenging situations as explained in the last paragraph of Section \ref{results}. Also, the training of the object recognition model can be made online to cope with the 
dynamic nature of the retail industry where new products appear continuously and replace the old ones. 

\section*{Acknowledgments} The authors would like to thank IIT(BHU), Varanasi for providing the necessary resources including servers, technicians, etc., to initiate research work in this area.
\bibliographystyle{unsrt}
\bibliography{refs}

\end{document}